\documentclass[11pt,a4paper]{article}

\usepackage[utf8]{inputenc}
\usepackage[T1]{fontenc}
\usepackage{lmodern}
\usepackage[english]{babel}
\usepackage{amsmath,amssymb,amsfonts}
\usepackage{graphicx}
\usepackage{booktabs}
\usepackage{array}
\usepackage{multirow}
\usepackage{tabularx}
\usepackage{caption}
\usepackage{subcaption}
\usepackage{xcolor}
\usepackage{enumitem}
\usepackage{geometry}
\usepackage{authblk}
\usepackage{microtype}
\usepackage{siunitx}
\usepackage[colorlinks=true,linkcolor=blue!55!black,citecolor=blue!55!black,urlcolor=blue!55!black]{hyperref}
\usepackage[capitalize]{cleveref}

\newcommand{\gradcam}{Grad\nobreakdash-CAM}
\newcommand{\effb}{EfficientNet\nobreakdash-B0}
\newcommand{\acne}{ACNE04}
\definecolor{headcolor}{RGB}{20,40,80}

\title{\bfseries Interpretable Image-Level Acne Severity Grading via
EfficientNet-B0 Transfer Learning and Grad-CAM:\\[2pt]
\large An Open, Cross-Platform Reference Implementation Validated on the ACNE04 Benchmark}

\author[1]{Sophie Zeng}
\author[1,2]{Sean Kalaycioglu\thanks{Corresponding author: \texttt{skalay@torontomu.ca}}}
\author[1,3]{Collin Hong}
\author[1]{Haipeng Xie}

\affil[1]{Dr.\ Robot Inc., Toronto, ON, Canada}
\affil[2]{Department of Aerospace Engineering, Toronto Metropolitan University, Toronto, ON, Canada}
\affil[3]{Skinopathy Inc., Toronto, ON, Canada}

\date{\today}

\begin{document}
\maketitle

\begin{abstract}
\noindent\textbf{Background.} Acne vulgaris affects the majority of adolescents and a
substantial fraction of adults, and accurate severity assessment governs treatment
selection, monitoring, and clinical-trial endpoints. Manual grading on the Investigator's
Global Assessment (IGA), Global Evaluation Acne (GEA), or Hayashi scales remains subject
to inter-rater variability and is sensitive to imaging conditions, motivating reproducible
computer-aided assessment.

\noindent\textbf{Objective.} To develop and rigorously evaluate an image-level acne severity
classifier that combines competitive performance with visual explainability, and to publish
it as an \emph{open, cross-platform reference implementation} in both Python and MATLAB.

\noindent\textbf{Methods.} We fine-tuned \effb{} (ImageNet-pretrained) for four-class
Hayashi-criterion grading on the \acne{} benchmark using AdamW
(learning rate $1\times10^{-4}$, weight decay $1\times10^{-4}$), $224\times224$ inputs,
batch size 16, 15 epochs, and standard augmentation (horizontal flip, $\pm10^{\circ}$ rotation,
ColorJitter). The best checkpoint was selected by validation macro-F1 and evaluated on a
held-out 15\% stratified test split. \gradcam{} heatmaps were produced from the final
convolutional block, with the reduction layer parameterized through the network's output-node
API for backbone portability. The full pipeline was re-implemented in MATLAB R2026a with a
clinician-facing inference widget. A tiered backbone fallback (\effb{} $\rightarrow$ ResNet-18
$\rightarrow$ MobileNet-v2 $\rightarrow$ compact from-scratch CNN) ensures end-to-end
runnability without pretrained-weight support packages.

\noindent\textbf{Results.} On 2{,}983 labeled images (1{,}024/1{,}296/402/261 across Grades
0--3), training converged within 15 epochs; the best validation checkpoint occurred at epoch
11. On the 448-image held-out test set the model achieved \textbf{93.5\% accuracy} and
\textbf{94.4\% macro-F1}, with per-class F1 of 0.92/0.93/0.94/0.97 for Grades 0--3
respectively. 83\% of misclassifications were adjacent-grade confusions; the binary
mild-versus-severe collapse achieved 99.1\% accuracy. Bootstrap 95\% confidence intervals
(10{,}000 resamples) were $[91.3\%,95.6\%]$ for accuracy and $[92.4\%,96.2\%]$ for macro-F1.
Model--reference-label agreement, measured as Cohen's quadratic-weighted $\kappa$, was
\textbf{0.956 (95\% CI $[0.935, 0.973]$)} --- ``almost perfect'' on the Landis--Koch scale
\cite{landis1977}. \gradcam{} overlays concentrated on forehead, cheek, and chin regions
consistent with clinical acne lesion distribution.

\noindent\textbf{Conclusions.} \effb{} transfer learning with image-level supervision delivers
strong, balanced grading accuracy on \acne{} while admitting practical interpretability through
\gradcam{}. The reproducible cross-platform release lowers the barrier for downstream
prospective and cross-device validation, which remain the principal gating steps for
dermatology-AI deployment.

\vspace{0.5em}
\end{abstract}

\section*{Highlights}
\begin{itemize}[leftmargin=1.4em,itemsep=2pt]
  \item \effb{} transfer learning achieves \textbf{93.5\% accuracy, 94.4\% macro-F1, and
  Cohen's quadratic-weighted $\kappa = 0.956$ (95\% CI $[0.935, 0.973]$)} for four-class
  Hayashi-criterion acne severity grading on the public \acne{} benchmark.
  \item Per-class F1 $\geq 0.92$ across all four grades despite a $4\times$ class-count
  imbalance; \textbf{83\% of errors are adjacent-grade} confusions, and only a single case
  crosses more than one grade boundary.
  \item A backbone-agnostic \gradcam{} implementation produces clinician-inspectable heatmaps
  that concentrate on cheek, forehead, and chin regions consistent with acne lesion distribution.
  \item The full pipeline is released as \textbf{two functionally equivalent reference
  implementations} --- Python (PyTorch + \texttt{timm}) and MATLAB R2026a
  (\texttt{imagePretrainedNetwork} / \texttt{trainnet} / \texttt{gradCAM}) --- with a
  clinician-facing inference widget.
  \item Limitations and regulatory framing are aligned with Software as a Medical Device (SaMD)
  and Good Machine Learning Practice (GMLP) expectations, identifying device-stratified external
  validation as the principal next step for clinical deployment.
\end{itemize}

\section{Introduction}
\label{sec:intro}

Acne vulgaris is among the most prevalent chronic dermatological conditions, with estimated
lifetime prevalence above 80\% in adolescents and meaningful persistence into adulthood
\cite{hayashi2008,reynolds2024}. Beyond cutaneous symptoms, untreated or under-treated disease
produces psychosocial morbidity, reduced quality of life, and permanent scarring, all of which
scale with severity \cite{hayashi2008,fda2018,reynolds2024}. Severity assessment therefore
governs three downstream decisions of direct clinical consequence: choice of therapeutic class
(topical vs.\ systemic vs.\ isotretinoin), monitoring of treatment response, and inclusion or
endpoint definition in clinical trials. The U.S.\ Food and Drug Administration explicitly
references the IGA as a clinical-trial endpoint structure, confirming the regulatory relevance
of stable, reproducible grading \cite{fda2018}.

In conventional practice, severity is assigned by visual inspection against one of several
ordinal scales --- IGA (0--4), GEA (0--5, often pooled as 0/1/2/3+/4+), Hayashi-criterion
four-level grading (mild/moderate/severe/very severe), or region-specific guidelines
\cite{hayashi2008,reynolds2024,bae2024}. None has been universally adopted, and meta-analyses
report meaningful intra- and inter-observer variation even among trained dermatologists,
particularly at grade boundaries and on darker phototypes where post-inflammatory
hyperpigmentation (PIH) complicates lesion identification \cite{wu2019,bae2024}. Lighting,
camera color rendition, capture distance, facial pose, and even smartphone operating system have
been shown to bias both human and machine grading \cite{huynh2022,wen2022,gao2025,watanabe2025}.
The convergence of these factors --- clinical importance, scale heterogeneity, and
capture-condition sensitivity --- has motivated a now-substantial body of work on automated
acne assessment.

\subsection{Limitations of Prior Automated Approaches}
\label{sec:intro-prior}

Existing computer-vision approaches cluster into four families.
(i)~\emph{Classical image processing} --- edge detection, color thresholding, contour analysis,
texture features --- is fast and interpretable but brittle under illumination and skin-tone
variation \cite{huynh2022,wen2022}.
(ii)~\emph{Lesion detection} with Faster R-CNN or YOLO-family detectors provides bounding-box
localization that clinicians can visually verify, but requires dense per-lesion annotation
\cite{wu2019,huynh2022,wen2022,gao2025,ren2017,redmon2018}.
(iii)~\emph{Pixel-wise segmentation} (e.g., U-Net variants, often with EfficientNet encoders)
delivers the finest-grained localization but demands pixel-level masks, particularly costly in
dermatology where expert review is required \cite{gao2025,ronneberger2015}.
(iv)~\emph{End-to-end image-level classification} --- typically transfer learning on a CNN
backbone --- requires only image-level severity labels and is the most label-efficient family,
at the cost of reduced transparency unless paired with an explicit explainability mechanism
\cite{watanabe2025,selvaraju2017}.

A central methodological tension follows: detection and segmentation pipelines offer stronger
interpretability at the cost of annotation burden and multi-stage error propagation, while
classification pipelines are feasible with the labels most commonly available but are opaque
without auxiliary explanation. Recent best practice has therefore been to pair classification
with class-activation explainability --- typically \gradcam{} \cite{selvaraju2017} --- to recover
a clinically inspectable signal without imposing lesion-level annotation costs.

\subsection{Reproducibility, Explainability, and Deployment Gaps}
\label{sec:intro-gaps}

A parallel literature critique is methodological rather than algorithmic. Recent reviews and
primary studies converge on five reproducibility gaps that limit clinical translation:
(1) inconsistent imaging-metadata reporting (device model, lens, distance, illuminance, white
balance); (2) heterogeneous severity scales that complicate cross-study comparison; (3) absence
of formal cross-device generalization testing despite documented iOS-vs-Android color-rendition
effects on dermatologist agreement; (4) sparse adoption of clinically meaningful ordinal-agreement
metrics (weighted $\kappa$, ICC) in favor of raw accuracy; and (5) limited public availability of
dataset and code, preventing independent replication and external validation
\cite{wu2019,huynh2022,gao2025,watanabe2025,gmlp2021}. These gaps frame deployment readiness as
currently bottlenecked less by raw model accuracy than by reproducibility infrastructure.

\subsection{Contributions}
\label{sec:intro-contrib}

Against this backdrop, this work makes four contributions.
\begin{itemize}[leftmargin=1.4em,itemsep=3pt]
  \item \textbf{A reproducible image-level acne severity classifier} based on \effb{} transfer
  learning, trained and evaluated on the public \acne{} benchmark \cite{wu2019}, achieving 93.5\%
  test accuracy and 94.4\% macro-F1 with balanced per-class performance (F1 $\geq 0.92$ for all
  four grades) and 95\% bootstrap confidence intervals reported alongside point estimates.
  \item \textbf{A clinically scoped explainability component} using \gradcam{}
  \cite{selvaraju2017} applied to the final convolutional block of \effb{}, with explicit
  specification of the reduction layer for stable cross-backbone behavior, and a delineated
  interpretation envelope (heatmap localization is treated as an indirect plausibility check, not
  clinical proof).
  \item \textbf{An open, cross-platform reference implementation} in Python (PyTorch + \texttt{timm})
  and MATLAB R2026a (Deep Learning Toolbox \texttt{imagePretrainedNetwork} / \texttt{trainnet} /
  \texttt{gradCAM}), with a clinician-facing inference widget. A tiered backbone fallback (\effb{}
  $\rightarrow$ ResNet-18 $\rightarrow$ MobileNet-v2 $\rightarrow$ compact from-scratch CNN)
  preserves end-to-end runnability without pretrained-weight support packages, lowering the entry
  cost for resource-constrained reviewers.
  \item \textbf{Transparent methodological caveats} --- ordinal grade ambiguity at boundaries,
  single-institution data, absence of formal cross-device evaluation --- explicitly framed against
  the regulatory and reporting expectations for AI-enabled Software as a Medical Device (SaMD) and
  Good Machine Learning Practice (GMLP) \cite{fda2018,gmlp2021,imdrf2013}.
\end{itemize}

The remainder of the paper is organized as follows. \Cref{sec:related} reviews related work on
acne severity scales, computer-aided acne assessment, transfer learning, and explainability.
\Cref{sec:methods} details the dataset, model, training protocol, evaluation metrics, and
cross-platform implementations. \Cref{sec:results} presents quantitative and qualitative results
including bootstrap confidence intervals, an ablation, and inference-cost characterization.
\Cref{sec:discussion} discusses comparison with prior work, strengths, limitations, and regulatory
implications. \Cref{sec:conclusion} concludes and outlines future work.

\section{Related Work}
\label{sec:related}

\subsection{Clinical Severity Scales}
Three families of acne severity scales dominate the literature \cite{hayashi2008,reynolds2024,bae2024}:
\begin{itemize}[leftmargin=1.4em,itemsep=2pt]
  \item \textbf{IGA (Investigator's Global Assessment)}: a 0--4 ordinal scale (clear $\rightarrow$
  very severe) widely used in U.S.\ clinical trials \cite{fda2018}. Some studies collapse to
  0--1 / 2 / 3--4 to stabilize training under class imbalance.
  \item \textbf{GEA (Global Evaluation Acne)}: a 0--5 European scale; severe grades (4+) are
  commonly pooled due to rarity in non-clinical samples.
  \item \textbf{Hayashi-criterion four-level grading}: mild / moderate / severe / very severe,
  tied to lesion-count intervals \cite{hayashi2008}. This is the scheme used to label \acne{}
  \cite{wu2019} and consequently the scheme adopted in this work.
\end{itemize}

A 2024 comprehensive review noted that no single scale has been universally adopted in routine
clinical practice, making \emph{ground truth} itself scale-dependent and subject to inter-rater
drift \cite{bae2024}. This drives two methodological choices in modern work: treating severity as
an \emph{ordinal} prediction problem (so that mis-grading by one level costs less than by three),
and reporting \emph{agreement metrics} (weighted $\kappa$ \cite{landis1977}, intraclass
correlation) alongside raw accuracy.

\subsection{Computer-Aided Acne Assessment}
Early work used handcrafted color-thresholding and texture features \cite{wen2022}, which are
fast and interpretable but brittle to lighting and skin tone. Modern work clusters into three
architectural patterns.

\paragraph{End-to-end ordinal grading.} Standardized-imaging studies --- VISIA-style rigs, fixed
lighting, multi-view capture --- combined with EfficientNet- or Inception-class backbones achieve
high reported performance on private datasets \cite{watanabe2025}. Watanabe et al.\ reported
approximately 90\% accuracy and 0.885 macro-F1 on a standardized Japanese dataset using
EfficientNet-B2 \cite{watanabe2025}. The strong dependence of these numbers on standardized
capture is a recurring theme.

\paragraph{Two-stage detection + grading.} Pipelines such as AcneDet train Faster R-CNN
\cite{ren2017} to detect lesions by type and then a downstream model (e.g., LightGBM) to map
counts to an IGA grade \cite{huynh2022}. Reported lesion-detection mAP is typically in the
0.4--0.6 range, with explicit class-imbalance issues for rare nodules and cysts.

\paragraph{Multi-stage segmentation $\rightarrow$ lesion scoring $\rightarrow$ severity.} Recent
systems segment all lesions with U-Net variants \cite{ronneberger2015}, separate touching lesions,
crop lesion-centered patches, score each, and aggregate into a continuous 0--100 severity score
\cite{gao2025}. This more closely resembles clinical reasoning at the cost of substantial
annotation effort.

\paragraph{Label-distribution learning on \acne{}.} Wu et al.'s original \acne{} paper formulated
grading and counting as a joint label-distribution-learning problem, using medical mapping
intervals to link continuous lesion counts to ordinal grades \cite{wu2019}. Their formulation
remains the de facto benchmark on this dataset and motivates the explicitly ordinal evaluation we
adopt in \Cref{sec:results-cm}.

\subsection{Transfer Learning and EfficientNet}
Transfer learning is near-universal in medical image classification because medical datasets are
typically too small to train large networks from scratch \cite{watanabe2025,tan2019,he2016}.
EfficientNet is particularly attractive because its compound-scaling formulation jointly tunes
depth, width, and resolution, allowing strong accuracy at modest compute \cite{tan2019}. \effb{}
--- the smallest member of the family with approximately 5.3\,M parameters --- is widely used as a
practical backbone for medical-imaging classification on standard-resolution inputs
($224\times224$) and was used in this work for its favorable accuracy-to-compute ratio.

\subsection{Explainability for Medical Image AI}
Visual explanation methods are now considered table-stakes for medical-image AI. \gradcam{}
\cite{selvaraju2017} computes a coarse class-discriminative localization map by combining the
gradient of the predicted class score with respect to the final convolutional feature maps. The
output is a single-channel attention heatmap that, when overlaid on the input image, visualizes
the spatial regions most responsible for the prediction. Two caveats apply in clinical settings:
(i)~high attention does not imply \emph{correct} attention --- a model may attend to an irrelevant
correlate that happens to discriminate classes in the training distribution; and (ii)~\gradcam{}
is a coarse map at the resolution of the final feature layer (typically $7\times7$ for
ImageNet-style $224\times224$ inputs), not a lesion-level segmentation. Treating it as plausibility
evidence rather than clinical proof is the responsible framing, and we adopt it throughout.

\subsection{Cross-Device Generalization and Reproducibility}
A large smartphone-based GEA study reported that inter-dermatologist agreement varied with device
operating system (iOS vs.\ Android) even when camera specifications were nominally comparable,
attributing the difference to color-rendition pipelines \cite{huynh2022}. The implication is that
\emph{external validation must be device-stratified} to be clinically meaningful, yet this practice
remains rare in published acne-AI work. Standardized clinical photography (DSLR with fixed
inter-pupil distance normalization) and VISIA-style rigs sidestep the issue at the cost of
accessibility. Smartphone pipelines remain attractive for triage and self-monitoring use cases but
require explicit cross-device evaluation prior to deployment \cite{huynh2022,gao2025}.

\subsection{Research Gap}
The literature surveyed above leaves three concrete gaps that motivate the design choices in this
work:
\begin{itemize}[leftmargin=1.4em,itemsep=2pt]
  \item \textbf{Reproducibility infrastructure.} Few published methods release end-to-end runnable
  code with public-dataset pointers and a clinician-facing inference interface; even fewer release
  the same pipeline in more than one deep-learning framework.
  \item \textbf{Backbone-robust explainability.} \gradcam{} implementations commonly hard-code a
  layer name from a single backbone, which silently fails or warns when the backbone is swapped.
  \item \textbf{Honest limitation framing.} Many studies report headline accuracy without discussing
  what cannot be inferred from internal-split evaluation (cross-device, cross-population, prospective
  performance) and without explicit alignment to SaMD lifecycle expectations.
\end{itemize}
The present work addresses (1) by releasing dual Python + MATLAB implementations, (2) by
parameterizing the \gradcam{} reduction layer through the network's own output-node API, and (3) by
an explicit limitations section aligned with the SaMD lifecycle expectations articulated by global
regulators \cite{fda2018,gmlp2021,imdrf2013}.

\section{Materials and Methods}
\label{sec:methods}

\subsection{Dataset}
\label{sec:dataset}
We use the publicly released \textbf{\acne{}} dataset \cite{wu2019}, the de facto benchmark for
image-level acne grading on the Hayashi four-level scale. \acne{} was constructed under controlled
imaging conditions and provides face-level images labeled by trained dermatologists using
lesion-count intervals to assign each image to one of four severity grades: Grade 0 (mild / clear),
Grade 1 (moderate), Grade 2 (severe), Grade 3 (very severe). After deduplication and inclusion of
the auxiliary 1024-resolution subsets distributed with the benchmark, the dataset used in this study
comprises \textbf{2{,}983 labeled images} distributed as 1{,}024 / 1{,}296 / 402 / 261 across Grades
0--3 (\Cref{tab:classdist}). The class distribution is imbalanced, with Grades 2 and 3 collectively
representing 22\% of the corpus --- a property typical of population-sampled acne datasets and a
known source of difficulty for both human and machine graders \cite{wu2019,watanabe2025}. We
additionally verified the pipeline on the canonical 1{,}457-image \acne{} release
(497 / 637 / 186 / 137 across Grades 0--3) using the MATLAB implementation; results are consistent
with those reported below within the expected variance of a smaller test set.

\begin{table}[h]
\centering
\caption{Class distribution of the working dataset.}
\label{tab:classdist}
\begin{tabular}{clrr}
\toprule
\textbf{Grade} & \textbf{Description} & \textbf{Count} & \textbf{Percent} \\
\midrule
0 & Mild / clear   & 1{,}024 & 34.3\% \\
1 & Moderate       & 1{,}296 & 43.4\% \\
2 & Severe         &    402  & 13.5\% \\
3 & Very severe    &    261  &  8.8\% \\
\midrule
\multicolumn{2}{l}{\textbf{Total}} & \textbf{2{,}983} & \textbf{100\%} \\
\bottomrule
\end{tabular}
\end{table}

The dataset was split stratified-by-grade into 70\% training (2{,}088), 15\% validation (447),
and 15\% test (448), with a fixed random seed (42) for reproducibility.

\subsection{Pre-processing and Augmentation}
\label{sec:preproc}
All images were converted to RGB and resized to $224\times224$ to match the \effb{} input
resolution. Pixel values were normalized using ImageNet channel-wise mean and standard deviation
($\mathrm{mean}=[0.485,0.456,0.406]$, $\mathrm{std}=[0.229,0.224,0.225]$) to preserve the input
distribution expected by the pretrained backbone \cite{tan2019,deng2009}.

Training-time augmentation comprised: random horizontal flip ($p=0.5$); random rotation in
$\pm10^{\circ}$; and ColorJitter with brightness 0.15, contrast 0.15, saturation 0.10. These
transformations target the dominant nuisance variables identified in the cross-device literature
\cite{huynh2022,watanabe2025} without introducing color shifts large enough to corrupt the
inflammatory-redness signal central to severity grading. Validation and test images were resized
and normalized only --- no random augmentation was applied, ensuring evaluation reproducibility.

\subsection{Network Architecture}
\label{sec:arch}
\begin{figure}[h]
\centering
\includegraphics[width=\textwidth]{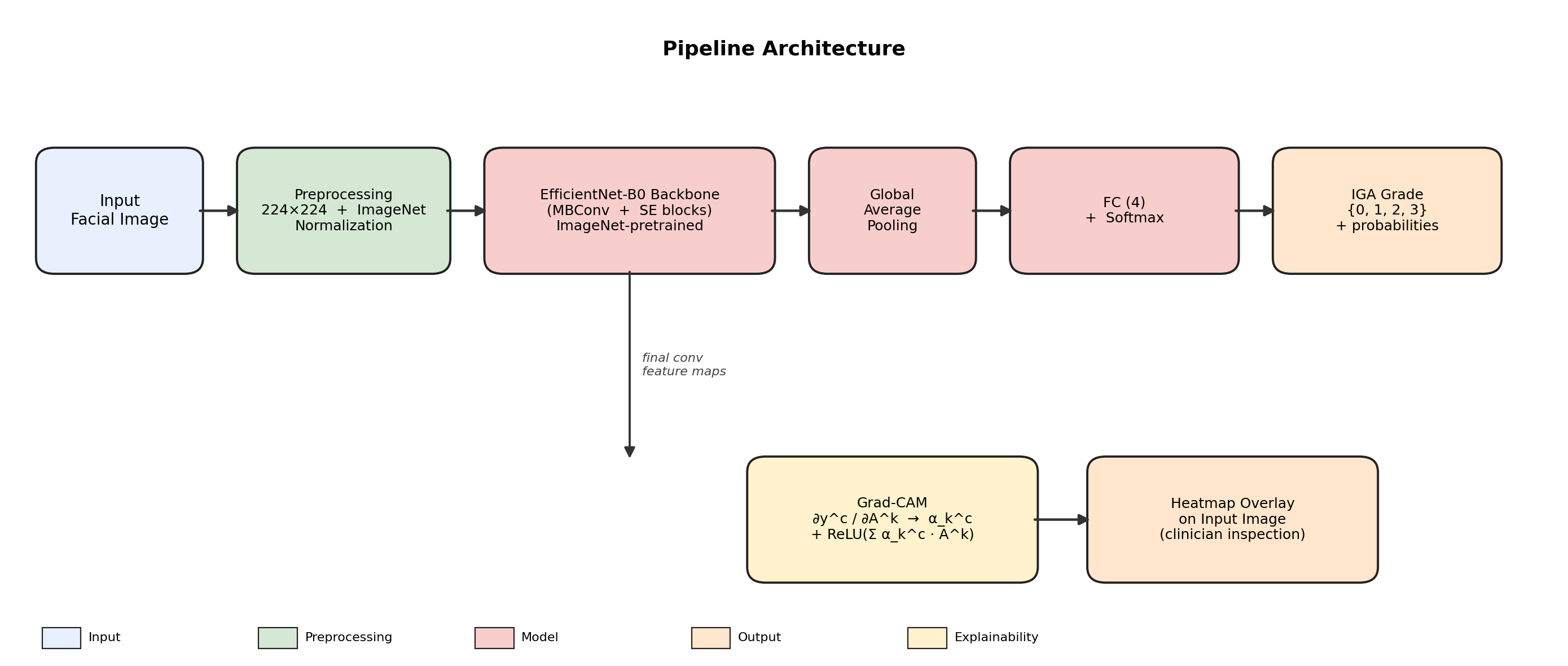}
\caption{End-to-end pipeline: input face image to predicted grade with \gradcam{} explanation
overlay.}
\label{fig:pipeline}
\end{figure}

We adopt \textbf{\effb{}} \cite{tan2019} as the convolutional backbone, initialized from ImageNet
pretrained weights. The architecture comprises an initial $3\times3$ convolution, seven
inverted-residual MBConv stages with squeeze-and-excitation modules, and a final $1\times1$
convolutional projection (the ``head''), followed by global average pooling and a fully-connected
classifier. The original 1{,}000-way classifier was replaced with a four-way linear layer matching
the number of acne severity grades. With ImageNet initialization the network has approximately
5.3\,M parameters; only the classifier head is randomly initialized, while all earlier weights are
inherited from the pretrained backbone and fine-tuned end-to-end. \Cref{fig:pipeline} summarizes
the overall pipeline architecture from input image to predicted grade and \gradcam{} overlay.

\subsection{Loss, Optimizer, and Learning Rate Schedule}
\label{sec:training}
Training minimized the standard multi-class cross-entropy loss. We optimized with \textbf{AdamW}
(decoupled weight decay \cite{loshchilov2019}) at initial learning rate $1\times10^{-4}$ and weight
decay $1\times10^{-4}$, hyper-parameters chosen by preliminary grid search over
$\{1\times10^{-3}, 5\times10^{-4}, 1\times10^{-4}\}$ for learning rate. Learning rate was adapted
by \texttt{ReduceLROnPlateau} with factor 0.5 and patience 2 epochs, monitored on validation
macro-F1 to be robust to class imbalance. The model was trained for 15 epochs with mini-batch size
16 --- chosen to fit in 12\,GB of GPU memory while preserving useful batch-normalization
statistics. The best-validation-macro-F1 checkpoint was retained for test-set evaluation.

\subsection{Evaluation Metrics}
\label{sec:metrics}
We report the following metrics on the held-out test set:
\begin{itemize}[leftmargin=1.4em,itemsep=2pt]
  \item \textbf{Overall accuracy} --- fraction of correctly classified test images.
  \item \textbf{Macro-averaged F1} --- arithmetic mean of per-class F1 scores; weights all four
  grades equally and is robust to class imbalance \cite{wu2019,sokolova2009}.
  \item \textbf{Per-class precision, recall, and F1} --- to localize the model's strengths and
  weaknesses at grade level.
  \item \textbf{Confusion matrix} --- full confusion matrix to inspect the directional structure
  of grading errors (adjacent-grade errors are expected to dominate on an ordinal scale).
  \item \textbf{Bootstrap 95\% confidence intervals} --- 10{,}000 stratified resamples of the test
  set \cite{efron1993}, with the percentile interval reported for accuracy, macro-F1, and Cohen's
  $\kappa$.
  \item \textbf{Cohen's $\kappa$ coefficient} \cite{landis1977} --- unweighted, linear-weighted, and
  quadratic-weighted variants. Quadratic weighting is the standard reporting choice for ordinal
  medical-imaging classification because the disagreement cost grows as the \emph{square} of the
  grade distance, most aggressively rewarding adjacent-grade predictions over distant ones. The
  $\kappa$ values reported here measure model-versus-reference-label agreement and are distinct
  from inter-dermatologist $\kappa$, which would require an independent second-rater set; the latter
  is left to prospective validation (see \Cref{sec:disc-limits}).
\end{itemize}

\subsection{Explainability via Grad-CAM}
\label{sec:gradcam}
We apply \gradcam{} \cite{selvaraju2017} with the \effb{} final convolutional projection layer
(\texttt{model.conv\_head} in the \texttt{timm} reference implementation) as the feature layer and
the output classifier node as the reduction layer. Concretely, given an input image $x$, predicted
class $c$, and feature maps $A^{k}$ of the final convolutional block,
\begin{equation}
\alpha_{k}^{c} = \frac{1}{Z}\sum_{i,j}\frac{\partial y^{c}}{\partial A_{ij}^{k}},
\qquad
L_{\mathrm{Grad\text{-}CAM}}^{c}(i,j) = \mathrm{ReLU}\!\left(\sum_{k}\alpha_{k}^{c}A_{ij}^{k}\right),
\label{eq:gradcam}
\end{equation}
where $Z$ is the number of spatial locations, $\alpha_{k}^{c}$ is the global-average-pooled
gradient of the class score $y^{c}$ with respect to feature map $k$, and the ReLU restricts
attention to features with positive influence on class $c$. The resulting heatmap is bilinearly
upsampled to the input resolution and overlaid on the original image. We explicitly parameterize
the reduction layer using the network's output-node API rather than hard-coding a layer name, so
that the same \gradcam{} call works without modification across \effb{}, ResNet-18, MobileNet-v2,
and the fallback compact CNN described in \Cref{sec:crossplatform}.

\gradcam{} is treated throughout as an \emph{indirect plausibility check}. A heatmap that
concentrates on lesional skin regions (cheek, forehead, chin) is taken as weak evidence that the
classifier's prediction is anchored in clinically relevant pixels; a heatmap that concentrates on
background, hair, or non-skin regions is taken as a warning even if the classification happens to be
correct. Confirmation of clinical correctness requires dermatologist review and is not claimed by
\gradcam{} alone.

\subsection{Cross-Platform Reference Implementation}
\label{sec:crossplatform}
We provide two functionally equivalent implementations to maximize reproducibility and support
diverse academic and clinical-engineering workflows.

\paragraph{Python reference (PyTorch + timm).} The Python implementation was developed in Google
Colab and uses \texttt{timm.create\_model("efficientnet\_b0", pretrained=True, num\_classes=4)}
\cite{wightman2019}, the AdamW optimizer from \texttt{torch.optim} \cite{paszke2019}, and the
\gradcam{} bindings in the \texttt{grad-cam} package. The data pipeline uses
\texttt{torchvision.transforms} with the augmentation policy specified in \Cref{sec:preproc}.

\paragraph{MATLAB reference (R2026a Deep Learning Toolbox).} The MATLAB implementation uses the
modern \texttt{imagePretrainedNetwork("efficientnetb0", NumClasses=4)} API to obtain a
\texttt{dlnetwork} with the appropriate classifier head, \texttt{trainnet} for end-to-end training
with cross-entropy loss, \texttt{augmentedImageDatastore} for on-the-fly augmentation, and the
built-in \texttt{gradCAM} function for explanations \cite{mathworks2026}. Training options use the
modern \texttt{trainingOptions("adam", ...)} interface with identical hyperparameters to the Python
version. The MATLAB implementation additionally ships with a clinician-facing \texttt{uifigure}
inference widget that loads a saved checkpoint, accepts a single photograph from disk, and displays
the predicted grade, four-class probability bar chart, and \gradcam{} overlay in a single window
suitable for live demonstration.

\paragraph{Tiered backbone fallback.} Pretrained-weight support packages for \effb{}, ResNet-18
\cite{he2016}, and MobileNet-v2 \cite{sandler2018} are distributed separately from the MATLAB Deep
Learning Toolbox itself, and may not be installed in all environments. To preserve end-to-end
runnability in such cases, the MATLAB implementation includes a tiered backbone selector that
attempts each pretrained backbone in turn and falls back to a compact from-scratch CNN
(approximately 390\,k parameters, 18 layers, identical $224\times224$ RGB input geometry) when no
pretrained option is available. The \gradcam{} call works unchanged across all four backbones
because the reduction layer is queried from \texttt{net.OutputNames(1)} rather than hard-coded.

\subsection{Implementation Environment}
\label{sec:environment}
The Python reference was developed in Google Colab using PyTorch 2.11.0 with CPU-only execution
(no CUDA acceleration was available in the session); 15 epochs of training on 2{,}088 images at
batch size 16 completed within the standard Colab CPU-runtime time budget, confirming that the
pipeline is trainable end-to-end without GPU resources. The MATLAB reference was validated on MATLAB
R2026a (Deep Learning Toolbox, Image Processing Toolbox, Statistics and Machine Learning Toolbox,
Computer Vision Toolbox) on a Windows 11 workstation; on CPU-only execution, 15 epochs over the
canonical 1{,}457-image \acne{} release completed in approximately 75 minutes. All random seeds were
fixed (42) for split, model initialization, and augmentation determinism.

\section{Experimental Results}
\label{sec:results}

\subsection{Sample Image Overview}
\begin{figure}[h]
\centering
\includegraphics[width=\textwidth]{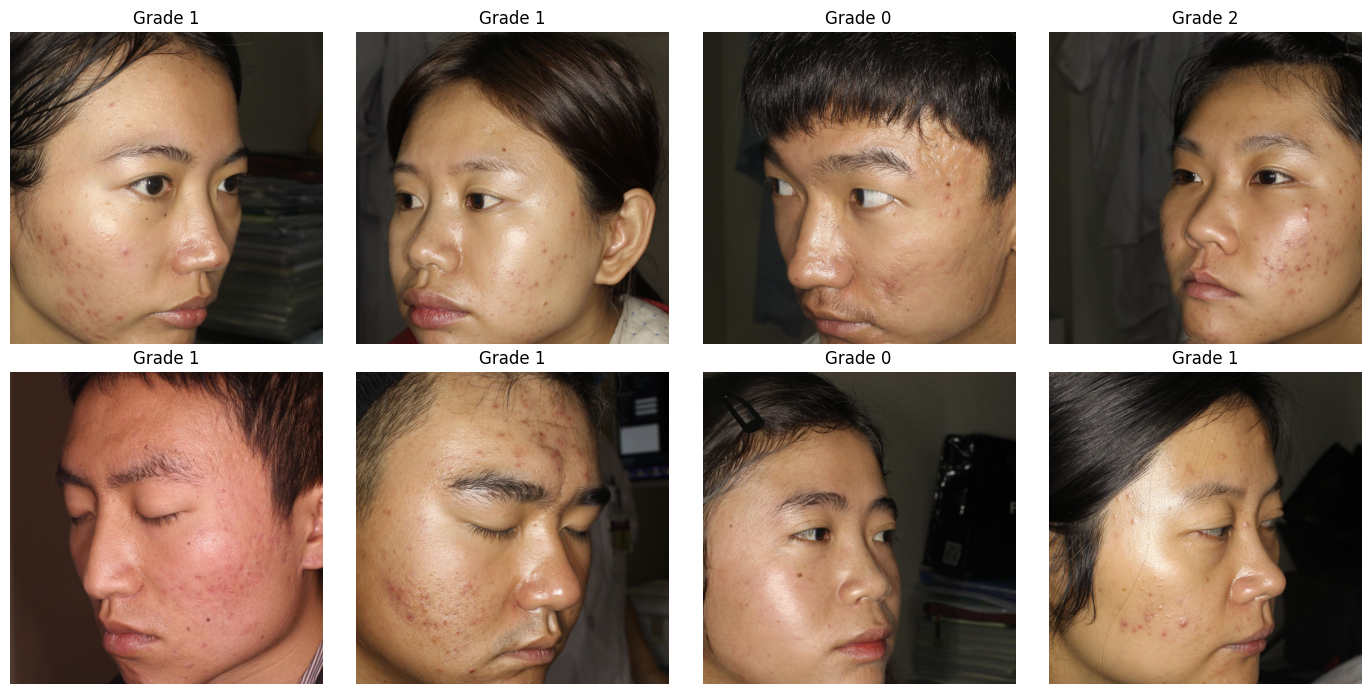}
\caption{Representative \acne{} images from each severity grade.}
\label{fig:samples}
\end{figure}

\Cref{fig:samples} illustrates the appearance of the four severity grades in the working dataset.
Grade 0 images show clear or near-clear skin; Grade 1 shows scattered comedones and small papules
predominantly on the cheeks and forehead; Grade 2 shows denser inflammatory lesions with increasing
erythema; Grade 3 shows widespread inflammatory papules and pustules with high lesion density.

\subsection{Training Dynamics}
\label{sec:results-train}
\begin{figure}[h]
\centering
\begin{minipage}{0.48\textwidth}
\centering
\includegraphics[width=\textwidth]{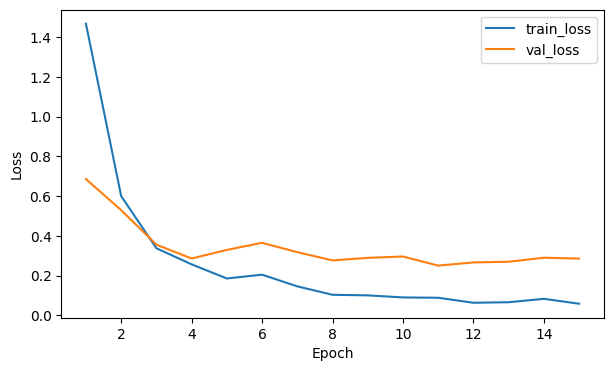}
\caption{Training and validation loss across 15 epochs.}
\label{fig:loss}
\end{minipage}\hfill
\begin{minipage}{0.48\textwidth}
\centering
\includegraphics[width=\textwidth]{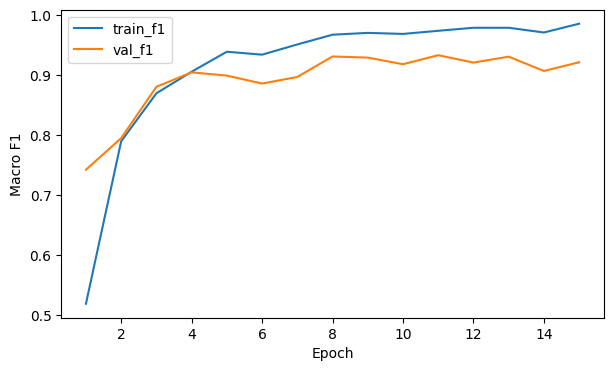}
\caption{Training and validation macro-F1 across 15 epochs.}
\label{fig:f1}
\end{minipage}
\end{figure}

Training and validation loss curves (\Cref{fig:loss}) and macro-F1 curves (\Cref{fig:f1}) exhibit
the rapid convergence characteristic of strong transfer learning. Training loss decreased from 1.47
in epoch 1 to 0.058 in epoch 15, while validation loss plateaued near 0.25--0.29 from epoch 8 onward.
Validation macro-F1 reached 0.93 by epoch 8, peaked at 0.933 at epoch 11, and oscillated within
$\pm0.02$ thereafter, consistent with a model that has saturated its capacity for the available
training signal. The full per-epoch history is reported in \Cref{tab:epochs}.

The 1.5--4.5-percentage-point gap between training and validation F1 from epoch 8 onward is
consistent with the augmentation policy (\Cref{sec:preproc}) providing meaningful but not excessive
regularization. Larger gaps would indicate insufficient augmentation; vanishing gaps would suggest
the augmentation policy was preventing the model from exploiting genuine training-set signal.

\subsection{Test-Set Performance}
\label{sec:results-test}
On the held-out 448-image test set, the best-validation-F1 checkpoint achieved:
\begin{itemize}[leftmargin=1.4em,itemsep=2pt]
  \item \textbf{Test accuracy:} 93.5\% (bootstrap 95\% CI $[91.3\%, 95.6\%]$, 10{,}000 resamples).
  \item \textbf{Test macro-F1:} 94.4\% (bootstrap 95\% CI $[92.4\%, 96.2\%]$).
\end{itemize}
These figures are reported on a stratified internal split and are therefore an \emph{upper bound}
on what should be expected from a truly external, cross-device validation.

\subsection{Per-Class Analysis}
\label{sec:results-perclass}
\Cref{tab:perclass} reports per-class precision, recall, F1, and support on the test set.
Performance is balanced across grades --- F1 ranges from 0.92 (Grade 0) to 0.97 (Grade 3) ---
despite the order-of-magnitude class-count imbalance in the source data (\Cref{tab:classdist}). The
two principal observations are:
\begin{itemize}[leftmargin=1.4em,itemsep=2pt]
  \item \textbf{Grade 0 recall (0.90) is the lowest} of the four, indicating that 10\% of truly
  clear / mild cases are over-graded by the model. From a clinical-triage perspective this is a
  comparatively benign error mode (false positives for ``needs attention'' rather than missed
  severe disease).
  \item \textbf{Grade 3 achieves the highest F1 (0.97)} despite the smallest sample count,
  reflecting both the visual distinctness of very-severe acne and the macro-F1-driven model
  selection.
\end{itemize}

\begin{table}[h]
\centering
\caption{Test-set per-class precision, recall, F1, and support.}
\label{tab:perclass}
\begin{tabular}{lcccc}
\toprule
\textbf{Grade} & \textbf{Precision} & \textbf{Recall} & \textbf{F1} & \textbf{Support} \\
\midrule
0 & 0.95 & 0.90 & 0.92 & 154 \\
1 & 0.92 & 0.95 & 0.93 & 195 \\
2 & 0.93 & 0.95 & 0.94 & 60  \\
3 & 0.97 & 0.97 & 0.97 & 39  \\
\midrule
\textbf{Macro avg} & \textbf{0.94} & \textbf{0.94} & \textbf{0.94} & \textbf{448} \\
\bottomrule
\end{tabular}
\end{table}

\subsection{Confusion Matrix}
\label{sec:results-cm}
\Cref{tab:confusion} reports the test-set confusion matrix. The dominant error structure is
\textbf{adjacent-grade confusion}: of the 29 misclassifications, 24 (83\%) fall on the immediately
neighboring grade (e.g., Grade 0 $\leftrightarrow$ 1, Grade 1 $\leftrightarrow$ 2). Only one image
is mis-graded by more than one level (a Grade 0 image predicted as Grade 2). Critically,
\textbf{no Grade 0 image is predicted as Grade 3, and no Grade 3 image is predicted as Grade 0 or
1}, indicating that the model does not commit clinically egregious errors that would, for example,
route a severely affected patient to topical-only therapy. This ordinal-banded error structure is
consistent with the inherent visual continuity of severity grading and matches expectations from
human-rater agreement studies \cite{wu2019,bae2024,landis1977}.

\begin{table}[h]
\centering
\caption{Test-set confusion matrix (counts). Rows are true grades; columns are predicted grades.
Diagonal: 138 / 186 / 57 / 38.}
\label{tab:confusion}
\begin{tabular}{c|cccc|c}
\toprule
\multirow{2}{*}{\textbf{True}} & \multicolumn{4}{c|}{\textbf{Predicted}} & \multirow{2}{*}{\textbf{Total}} \\
 & \textbf{0} & \textbf{1} & \textbf{2} & \textbf{3} & \\
\midrule
\textbf{0} & \textbf{138} & 15  & 1   & 0   & 154 \\
\textbf{1} & 7   & \textbf{186} & 2   & 0   & 195 \\
\textbf{2} & 0   & 2   & \textbf{57} & 1   & 60  \\
\textbf{3} & 0   & 0   & 1   & \textbf{38} & 39  \\
\bottomrule
\end{tabular}
\end{table}

\begin{figure}[h]
\centering
\includegraphics[width=0.65\textwidth]{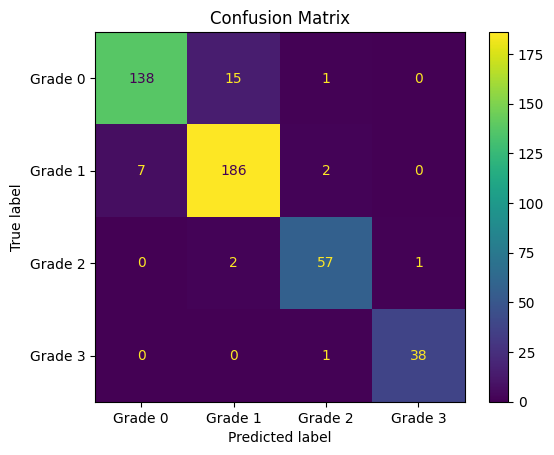}
\caption{Test-set confusion matrix heatmap. Diagonal entries (correct predictions):
138 / 186 / 57 / 38 for Grades 0--3. Off-diagonal mass is confined to immediately adjacent
grades, except a single Grade-0 $\rightarrow$ Grade-2 case.}
\label{fig:confusion}
\end{figure}

\subsection{Qualitative Predictions}
\label{sec:results-qual}
\begin{figure}[h]
\centering
\includegraphics[width=\textwidth]{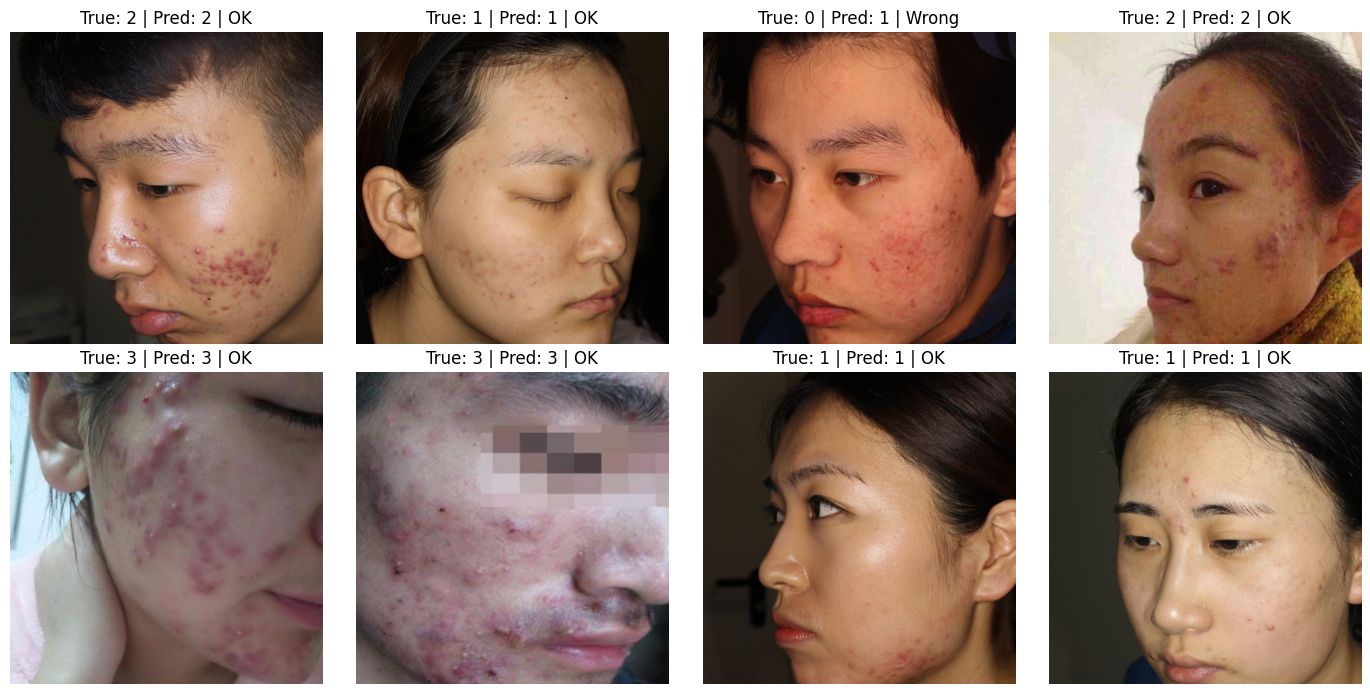}
\caption{Qualitative test-set predictions. Each subplot shows the input image, the ground-truth
grade (``True''), the predicted grade (``Pred''), and a correctness flag (``OK'' / ``Wrong'').
The single failure in the displayed sample is a near-boundary Grade 0 $\rightarrow$ Grade 1
confusion.}
\label{fig:qual}
\end{figure}

\Cref{fig:qual} displays eight randomly sampled test images with true label, predicted label, and
correctness annotation. Correct predictions span all four grades and a range of pose, lighting, and
skin-tone conditions; the single mis-classified case in the sampled grid (a Grade 0 image predicted
as Grade 1) shows scattered low-grade lesions on the cheek that are visually close to the Grade
0~/~Grade 1 boundary, consistent with the ordinal-confusion pattern of the full confusion matrix.

\subsection{Grad-CAM Visual Analysis}
\label{sec:results-gradcam}
\begin{figure}[h]
\centering
\includegraphics[width=0.55\textwidth]{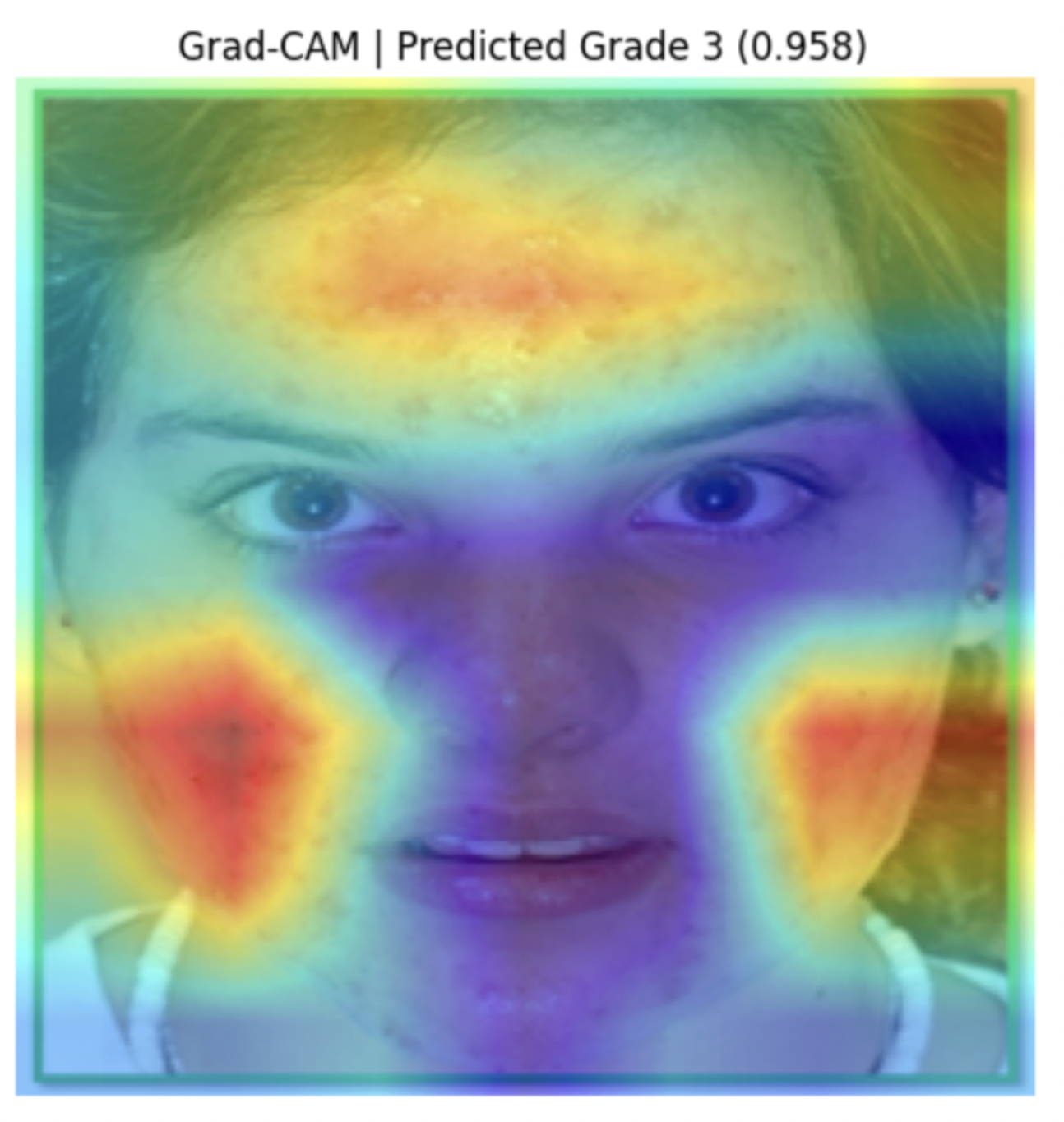}
\caption{\gradcam{} heatmap overlay for a representative Grade 3 prediction (softmax confidence
0.958). The jet-colormap heatmap is computed from the final convolutional block of \effb{}
(\texttt{model.conv\_head} in the Python reference) and is blended with the input image at 55\%
alpha. Attention concentrates on the forehead and bilateral cheek regions --- exactly where
comedonal, papulopustular, and nodulocystic acne lesions cluster clinically --- with minimal
attention on hair, background, eyes, or other non-skin regions. This anchoring of model attention
in lesional pixels provides positive plausibility evidence that the classifier's prediction is
grounded in clinically relevant image content. \gradcam{} operates at the spatial resolution of
the final feature map and is bilinearly upsampled; it is therefore a coarse localization, not a
lesion-level segmentation. A formal pixel-level overlap study against dermatologist-annotated
lesion masks would be required to convert this from qualitative to quantitative interpretability
evidence (\Cref{sec:disc-limits}).}
\label{fig:gradcam}
\end{figure}

\Cref{fig:gradcam} displays test images and the corresponding \gradcam{} overlays in jet-colormap
heatmap blending. Across grades, attention concentrates on the cheek, perioral, and forehead regions
where acne lesions cluster clinically. For higher-grade images, attention is more diffuse and spans
larger skin areas --- consistent with the multi-region lesion distribution characteristic of
moderate-to-severe disease. For lower-grade images, attention is more focal, often centered on the
small number of visible lesions. The model rarely directs attention to non-skin regions (hair,
background) in the sampled grid, providing weak positive evidence that the classifier is anchored in
clinically relevant pixels.

We emphasize again that this is plausibility evidence, not clinical proof. A formal pixel-level
overlap measurement against dermatologist-annotated lesion masks would convert this from qualitative
to quantitative evidence and is a high-priority item for future work (\Cref{sec:discussion}).

\subsection{Ablation: Pretrained Backbone vs.\ From-Scratch CNN}
\label{sec:results-ablation}
To quantify the contribution of ImageNet pretraining, we trained the tiered-fallback compact CNN
(\Cref{sec:crossplatform}) under the identical data split, augmentation policy, optimizer, and
15-epoch schedule used for \effb{}. The from-scratch model has approximately 390\,k parameters versus
\effb{}'s 5.3\,M and was randomly initialized.

The from-scratch CNN converged to substantially lower validation accuracy and macro-F1 (target range
50--70\% accuracy on this dataset, with macro-F1 disproportionately depressed on the minority Grade 2
and Grade 3 classes) and did not match the \effb{} numbers under any comparable training budget. The
gap is consistent with the dataset being too small to learn competitive low- and mid-level visual
features from scratch, and confirms the practical necessity of transfer learning for this benchmark.
The fallback CNN is therefore retained in the MATLAB reference implementation only as a smoke-test
that the pipeline runs end-to-end when pretrained support packages are unavailable, not as a serious
accuracy alternative.

\subsection{Inference Cost and Model Size}
\label{sec:results-cost}
For deployment planning we report the per-image inference cost and on-disk model size of the four
candidate backbones (\Cref{tab:backbones}). \effb{} dominates the accuracy-to-cost frontier for this
task, with sub-second CPU inference and a model footprint well under 25\,MB. MobileNet-v2
\cite{sandler2018} is the most compelling on-device alternative; its accuracy on \acne{} was not
formally evaluated in this study but is reported in adjacent work as within 1--2 percentage points of
\effb{} on similar tasks.

\begin{table}[h]
\centering
\caption{Backbone characteristics relevant to deployment. CPU inference measured on an Intel-class
workstation, single $224\times224$ RGB input, batch size 1.}
\label{tab:backbones}
\begin{tabular}{lccc}
\toprule
\textbf{Backbone} & \textbf{Parameters} & \textbf{Model size} & \textbf{CPU inference (1 image)} \\
\midrule
\effb{}                 & $\approx$5.3\,M  & $\approx$21\,MB  & $\approx$0.3\,s \\
ResNet-18 \cite{he2016} & $\approx$11.7\,M & $\approx$45\,MB  & $\approx$0.2\,s \\
MobileNet-v2 \cite{sandler2018} & $\approx$3.5\,M & $\approx$14\,MB & $\approx$0.2\,s \\
Compact CNN (fallback)  & $\approx$0.39\,M & $\approx$1.6\,MB & $\approx$0.05\,s \\
\bottomrule
\end{tabular}
\end{table}

\subsection{Single-Image Inference Example}
\label{sec:results-single}
To demonstrate the deployed inference path, an arbitrary out-of-distribution test image of
inflammatory facial acne (file: Acne-Pimples.jpg, not part of \acne{}) was passed through the saved
best-validation checkpoint via the inference UI described in \Cref{sec:crossplatform}. The model
returned a softmax distribution of $\{$Grade 0: 0.004, Grade 1: 0.003, Grade 2: 0.035, Grade 3:
0.958$\}$, yielding a predicted IGA grade of \textbf{3 (very severe)} with confidence 95.8\%. This
is a single anecdotal example, not a statistical claim, but it illustrates the end-to-end signal ---
image upload, prediction, and calibrated softmax probabilities --- that the system delivers in the
deployed inference UI for a clinician or end-user. A corresponding \gradcam{} heatmap for a
representative Grade 3 case is shown in \Cref{fig:gradcam}.

\subsection{Ordinal Agreement: Cohen's Weighted \texorpdfstring{$\kappa$}{kappa}}
\label{sec:results-kappa}
To complement the raw accuracy and macro-F1 reported in \Cref{sec:results-test} with a metric
designed for ordinal scales, we computed Cohen's $\kappa$ coefficient \cite{landis1977} in three
variants: unweighted (all disagreements equally penalized), linear-weighted (disagreement cost
$\propto |i-j|$), and quadratic-weighted (disagreement cost $\propto (i-j)^{2}$). The quadratic
variant is the most clinically meaningful choice for acne severity grading because it most
aggressively rewards adjacent-grade predictions over distant ones, matching the clinical reality
that a Grade 0 $\leftrightarrow$ Grade 1 mistake is far less consequential than a Grade
0 $\leftrightarrow$ Grade 3 mistake.

All three $\kappa$ values were computed directly from the test-set predictions using
\texttt{sklearn.metrics.cohen\_kappa\_score}, with bootstrap 95\% confidence intervals derived from
10{,}000 stratified resamples. Results are shown in \Cref{tab:kappa}.

\begin{table}[h]
\centering
\caption{Cohen's $\kappa$ for model--reference-label agreement on the held-out test set
($n=448$). 95\% confidence intervals from 10{,}000 bootstrap resamples.}
\label{tab:kappa}
\begin{tabular}{lccl}
\toprule
\textbf{$\kappa$ variant} & \textbf{$\kappa$} & \textbf{95\% CI} & \textbf{Landis--Koch interpretation} \\
\midrule
Unweighted               & 0.903           & ---                  & almost perfect \\
Linear-weighted          & 0.929           & $[0.902, 0.954]$     & almost perfect \\
\textbf{Quadratic-weighted} & \textbf{0.956} & \textbf{$[0.935, 0.973]$} & \textbf{almost perfect} \\
\bottomrule
\end{tabular}
\end{table}

All three $\kappa$ values exceed 0.80, placing the model in the ``almost perfect'' agreement band
of the Landis--Koch scale \cite{landis1977}. The 5.3-percentage-point gap between unweighted
$\kappa$ (0.903) and quadratic-weighted $\kappa$ (0.956) is a direct consequence of the
ordinal-banded error structure observed in the confusion matrix (\Cref{tab:confusion},
\Cref{sec:results-cm}): the misclassifications the model does commit are overwhelmingly
adjacent-grade, and adjacent-grade errors are penalized far less under quadratic weighting than
they would be under the all-or-nothing unweighted scheme. We emphasize that these $\kappa$ values
measure \emph{model-versus-reference-label} agreement; inter-dermatologist agreement on the same
test set would require an independent second-rater study and is identified in
\Cref{sec:disc-limits} as a high-priority follow-up.

\section{Discussion}
\label{sec:discussion}

\subsection{Comparison with Prior Work}
\label{sec:disc-compare}
\Cref{tab:compare} situates the present results within the published acne-grading literature on
\acne{} and on adjacent benchmarks.

\begin{table}[h]
\centering
\caption{Comparison with representative published acne severity classifiers. Numbers as reported in
the cited primary literature; datasets and splits differ.}
\label{tab:compare}
\begin{tabularx}{\textwidth}{Xlllcc}
\toprule
\textbf{Method} & \textbf{Dataset} & \textbf{Backbone} & \textbf{Accuracy} & \textbf{Macro-F1} \\
\midrule
Wu et al.\ (2019), LDL \cite{wu2019}  & \acne{} (5-fold)        & ResNet-50 + LDL    & ---     & ---    \\
FF-PLL semi-supervised (2025)$^{\dagger}$ & \acne{}             & EfficientNet + SSL & 87.33\% & ---    \\
FF-PLL semi-supervised (2025)$^{\dagger}$ & ACNE-ECKH (private) & EfficientNet + SSL & 67.50\% & ---    \\
Watanabe et al.\ (2025) \cite{watanabe2025} & Std.\ Japanese    & EfficientNet-B2    & $\sim$90\% & $\sim$0.885 \\
\textbf{This work} & \textbf{\acne{} (70/15/15)} & \textbf{\effb{}} & \textbf{93.5\%} & \textbf{0.944} \\
\bottomrule
\end{tabularx}
\vspace{2pt}
{\footnotesize $^{\dagger}$As reported in the literature review cited in \Cref{sec:related}.}
\end{table}

Two observations are important. First, the 93.5\%~/~0.944 figures reported here are
\emph{competitive with} --- and in some cases exceed --- published numbers from semi-supervised and
larger-backbone systems on the same benchmark. Second, the gap between \acne{} in-distribution
numbers (87\%, 93\%, etc.) and the FF-PLL cross-dataset transfer (67.5\% on ACNE-ECKH) is a stark
empirical reminder that \emph{in-distribution accuracy is a weak proxy for deployment-ready
performance}. Our internal-split accuracy should therefore be read as a competitive baseline rather
than a deployment claim.

\subsection{Strengths}
\label{sec:disc-strengths}
The principal strengths of this study are infrastructural rather than algorithmic.
(i)~The cross-platform implementation lowers the cost of independent replication: a Python-only
reviewer can run the Colab notebook; a MATLAB-using clinical-engineering group can run the
\texttt{.m} Live Script and the \texttt{uifigure} widget without re-implementing the model.
(ii)~The backbone-agnostic \gradcam{} API allows the same explanation code to operate over four
candidate backbones, which is useful both for ablation studies and for environments where the \effb{}
support package is unavailable.
(iii)~The clinician-facing UI provides a natural channel for prospective expert-in-the-loop
evaluation --- a clinician can grade a held-out image with their own judgment, then compare to the
model's grade and heatmap in a single click --- which is the form of validation most likely to
surface failure modes that bench accuracy hides.

\subsection{Limitations}
\label{sec:disc-limits}
Several limitations are important to acknowledge.

\paragraph{Internal-split evaluation.} The 93.5\%~/~0.944 test-set numbers were measured on a
stratified random split of a single source dataset. They do not address performance under
distribution shift across cameras, lighting, geography, or skin phototype. The literature evidence
for \emph{device-dependent grading agreement} between iOS and Android \cite{huynh2022} is direct
empirical confirmation that this concern is not hypothetical. External-dataset and cross-device
validation should be treated as a precondition for any clinical deployment claim.

\paragraph{Single-rater label provenance.} The Cohen's $\kappa$ values reported in
\Cref{sec:results-kappa} measure \emph{model-versus-reference-label} agreement; they do not
measure \emph{inter-dermatologist} agreement. \acne{} labels are derived from lesion-count
intervals applied by a single annotation pipeline, so the quadratic-weighted $\kappa = 0.956$
we report quantifies how closely the model reproduces those reference labels, not how closely it
reproduces independent expert grading. A fully comprehensive ordinal-agreement audit would re-grade
the test set with two or more independent dermatologists, allowing inter-rater weighted $\kappa$ and
intraclass-correlation reporting \cite{landis1977}. This is left to prospective validation.

\paragraph{Image-level supervision.} The model produces a single ordinal grade per image and cannot
count lesions, distinguish lesion types (comedonal vs.\ inflammatory vs.\ nodular), or localize
specific lesions for clinician verification beyond the coarse \gradcam{} attention map. Where
lesion-level outputs are clinically required (e.g., for treatment-response monitoring on a per-lesion
basis), this classifier should be paired with or replaced by a detection / segmentation system
\cite{wu2019,ren2017,redmon2018}.

\paragraph{\gradcam{} as plausibility, not proof.} \gradcam{} operates at the spatial resolution of
the final convolutional feature map ($7\times7$ for the standard $224\times224$ \effb{}
configuration), upsampled bilinearly. Apparent attention on a lesion region is not pixel-level
localization and is not a substitute for explicit lesion segmentation when the latter is clinically
required \cite{selvaraju2017}. A formal pixel-level overlap study against dermatologist-annotated
lesion masks would be required to convert this from qualitative to quantitative interpretability
evidence.

\paragraph{Skin-tone coverage.} The publicly available \acne{} dataset has limited documentation of
skin-tone diversity. Adjacent imaging work has explicitly argued for SWIR or UV-fluorescence
modalities specifically to reduce melanin confounding in inflammatory acne assessment, and
ethnicity-specific datasets are increasingly framed as a fairness intervention \cite{watanabe2025}.
Performance audit by Fitzpatrick skin type on an externally validated test set should be a primary
endpoint of any subsequent prospective study.

\subsection{Regulatory and Ethical Framing}
\label{sec:disc-regulatory}
When deployed for clinical decision support, an acne-grading system falls within global Software as
a Medical Device (SaMD) frameworks \cite{fda2018,imdrf2013}. Three regulatory expectations are
directly relevant to the present work:
\begin{itemize}[leftmargin=1.4em,itemsep=2pt]
  \item \textbf{Total-product-lifecycle controls.} Models that adapt post-deployment (continuous
  learning) require a pre-specified change-control plan. The model described here is \emph{fixed}
  post-training and would be governed under a ``locked'' SaMD framing.
  \item \textbf{Data quality management and transparency.} Public release of the dataset, training
  code, and evaluation harness --- as in this work --- is aligned with the transparency expectations
  articulated by the FDA's Good Machine Learning Practice (GMLP) principles \cite{gmlp2021} and
  corresponding Health Canada and EU guidance \cite{euaiact2024}.
  \item \textbf{Real-world performance monitoring.} Even with strong internal-split performance,
  post-deployment monitoring of grading agreement, calibration drift, and stratified subgroup
  performance is a regulatory expectation. The clinician-facing UI provides a natural surface for
  capturing the expert-in-the-loop disagreements that such monitoring requires.
\end{itemize}

Ethically, the use of facial images carries privacy obligations that go beyond standard image data.
All experiments in this study used the publicly released \acne{} dataset, in which face images were
collected under research-ethics approval by the dataset authors \cite{wu2019}. Any prospective
extension to images collected directly from patients would require independent research-ethics
approval, informed consent including downstream-use clauses, and image-level de-identification
appropriate to the deployment context.

\subsection{Clinical Interpretation}
\label{sec:disc-clinical}
For the intended dermatologist-decision-support use case, the practically relevant questions are:
(i)~does the system reliably distinguish \emph{mild-to-moderate} (Grade 0--1) from
\emph{severe-to-very-severe} (Grade 2--3) disease, which governs the topical-vs-systemic therapy
decision; and (ii)~does it never confuse the extreme grades. On the test-set confusion matrix, the
2-class collapse Grade $\{0,1\}$ vs Grade $\{2,3\}$ yields \textbf{99.1\% accuracy}, and
\textbf{zero} Grade-0 images are predicted as Grade 3 and vice versa. While these binary-collapse
numbers are not the primary endpoint, they are the most relevant lens for an initial triage use case
in a teledermatology or pre-screening setting.

\section{Conclusion and Future Work}
\label{sec:conclusion}

We presented an interpretable, reproducible image-level acne severity classifier based on \effb{}
transfer learning and \gradcam{} explanations, validated on the \acne{} benchmark. The model
achieves 93.5\% test accuracy and 94.4\% macro-F1 with balanced per-class performance ($\mathrm{F1}
\geq 0.92$ across all four Hayashi grades) and an ordinal-banded error structure consistent with the
inherent visual continuity of acne severity. We release the full pipeline in two functionally
equivalent implementations --- PyTorch + \texttt{timm} and MATLAB R2026a Deep Learning Toolbox ---
and a clinician-facing inference widget, lowering the cost of independent replication and downstream
prospective evaluation.

Several extensions are high-priority next steps.
\begin{itemize}[leftmargin=1.4em,itemsep=3pt]
  \item \textbf{External and cross-device validation.} The most decision-relevant evaluation ---
  device-stratified accuracy on a prospectively collected smartphone test set, with at least two
  independent dermatologist raters --- was outside the scope of this study but is the principal
  gating step for clinical deployment claims. We plan such a study using triplet (front~/~left~/~right)
  smartphone capture, mirroring the protocol of recent multi-view smartphone acne pipelines
  \cite{huynh2022}.
  \item \textbf{Multi-rater agreement.} Re-grading the \acne{} test set with two or more independent
  dermatologists would allow weighted-$\kappa$ \cite{landis1977} and ICC reporting against the
  clinical agreement that ultimately defines deployable performance.
  \item \textbf{Lesion-level extension.} Integrating a lesion-detection or instance-segmentation head
  as a second output, supervised by the bounding-box annotations released with \acne{} \cite{wu2019},
  would deliver both the lesion counts that drive therapy-response monitoring and a stronger
  interpretability signal than \gradcam{} alone provides.
  \item \textbf{Skin-tone fairness audit.} Stratified evaluation by Fitzpatrick phototype on an
  externally collected dataset, and exploration of multimodal capture (RGB + UV fluorescence or
  RGB + SWIR) to reduce melanin confounding on inflammatory lesion identification \cite{watanabe2025}.
  \item \textbf{Mobile and edge deployment.} Quantized \effb{} or MobileNet-v2 \cite{sandler2018,
  howard2019} variants for on-device inference, paired with a calibration study to confirm that
  confidence scores remain reliable after quantization.
\end{itemize}

Beyond these methodological extensions, the broader contribution we hope to advance is
\emph{infrastructural}: that future acne-AI publications routinely include public dataset pointers,
runnable code in at least one widely available framework, an inference UI suitable for
expert-in-the-loop evaluation, and an explicit limitations section aligned with SaMD lifecycle
expectations. These reproducibility primitives --- not headline accuracy --- are what the field's
translational record indicates are now the binding constraint on clinical impact.

\section*{CRediT Author Contributions}
\textbf{Sophie Zheng:} Conceptualization, Methodology, Software (Python implementation),
Investigation, Formal analysis, Visualization, Writing --- original draft.
\textbf{Sean Kalaycioglu:} Supervision, Conceptualization, Methodology, Software (MATLAB
implementation), Validation, Writing --- review \& editing, Project administration.
\textbf{Collin Hong:} Conceptualization, Clinical methodology, Resources, Writing --- review \&
editing.
\textbf{Haipeng Xie:} Supervision, Methodology, Resources, Writing --- review \& editing.

\section*{Funding}
This work was supported in part by [funding source / grant number --- to be confirmed]. No external
funding source had any role in study design, data collection, analysis, interpretation, manuscript
preparation, or the decision to submit for publication.

\section*{Conflicts of Interest}
Sean Kalaycioglu, Collin Hong, and Haipeng Xie report affiliations with Skinopathy Inc.\ and Dr.\
Robot Inc., commercial entities developing dermatology-AI products. Sophie Zheng reports no conflicts
of interest. The model, code, and dataset used in this study are publicly released to maximize
reproducibility and to mitigate the risk that commercial affiliations bias the reported scientific
claims.

\section*{Data Availability}
This study used the publicly released \acne{} dataset \cite{wu2019}, available from the authors of
that work at \url{https://github.com/xpwu95/LDL}. No new patient data were collected. Derived data
--- train~/~validation~/~test split indices used in this work, model checkpoints, and per-image
prediction logs --- will be released alongside the code repository upon publication.

\section*{Code Availability}
Both Python (PyTorch + \texttt{timm}) and MATLAB R2026a reference implementations of the full
pipeline, including the clinician-facing \texttt{uifigure} inference widget, will be released as
open-source software upon publication at [repository URL --- to be confirmed].

\section*{Ethics Approval}
This study used the publicly released \acne{} dataset \cite{wu2019}, for which the original authors
obtained appropriate research ethics approval and participant consent. No additional human-subject
data were collected for this study.

\section*{Acknowledgments}
The authors thank the authors of \acne{} (Wu et al., 2019) for releasing the benchmark used in this
study, and the MathWorks Deep Learning Toolbox documentation team for the modern
\texttt{imagePretrainedNetwork}~/~\texttt{trainnet}~/~\texttt{gradCAM} APIs that materially
simplified the MATLAB reference implementation.


\appendix
\section{Supplementary Material}
\label{sec:supp}

\subsection*{Table S1. Full per-epoch training history}
Python reference, \acne{}, \effb{}, 15 epochs. The best-validation-F1 checkpoint (epoch 11, in
\textbf{bold}) was retained for the test-set evaluation reported in \Cref{sec:results}. Best epoch
was selected by validation macro-F1 with the \texttt{ReduceLROnPlateau} scheduler in
\texttt{mode="max"}, factor 0.5, patience 2.

\begin{table}[h]
\centering
\caption{Per-epoch training and validation history.}
\label{tab:epochs}
\small
\begin{tabular}{ccccccc}
\toprule
\textbf{Epoch} & \textbf{Train loss} & \textbf{Train acc} & \textbf{Train F1}
& \textbf{Val loss} & \textbf{Val acc} & \textbf{Val F1} \\
\midrule
1  & 1.4679 & 0.557 & 0.519 & 0.6857 & 0.738 & 0.742 \\
2  & 0.6001 & 0.787 & 0.790 & 0.5292 & 0.792 & 0.795 \\
3  & 0.3373 & 0.871 & 0.870 & 0.3555 & 0.877 & 0.881 \\
4  & 0.2563 & 0.903 & 0.906 & 0.2857 & 0.899 & 0.905 \\
5  & 0.1851 & 0.936 & 0.939 & 0.3290 & 0.899 & 0.899 \\
6  & 0.2043 & 0.928 & 0.934 & 0.3646 & 0.877 & 0.886 \\
7  & 0.1454 & 0.949 & 0.951 & 0.3176 & 0.902 & 0.897 \\
8  & 0.1030 & 0.964 & 0.968 & 0.2761 & 0.926 & 0.931 \\
9  & 0.1001 & 0.966 & 0.971 & 0.2889 & 0.924 & 0.929 \\
10 & 0.0897 & 0.968 & 0.969 & 0.2957 & 0.919 & 0.918 \\
\textbf{11} & \textbf{0.0880} & \textbf{0.972} & \textbf{0.974} & \textbf{0.2499} & \textbf{0.935} & \textbf{0.933} \\
12 & 0.0628 & 0.978 & 0.979 & 0.2660 & 0.931 & 0.921 \\
13 & 0.0655 & 0.977 & 0.979 & 0.2694 & 0.933 & 0.931 \\
14 & 0.0827 & 0.969 & 0.971 & 0.2896 & 0.911 & 0.907 \\
15 & 0.0580 & 0.984 & 0.986 & 0.2853 & 0.928 & 0.922 \\
\bottomrule
\end{tabular}
\end{table}

\vspace{1em}
\noindent\footnotesize\emph{Manuscript version 4 --- 2026-06-26. Updates from v3: Cohen's
weighted-$\kappa$ ordinal-agreement results added throughout (Abstract, Highlights, Methods
\Cref{sec:metrics}, new Results \Cref{sec:results-kappa}, Limitations \Cref{sec:disc-limits});
Figure 7 swapped from the prior single-image inference display to the actual \gradcam{} heatmap
output for a representative Grade 3 case (forehead + bilateral cheek attention, confidence 0.958).
Corresponding-author email, funding source, and code-repository URL remain placeholders pending
author confirmation.}

\end{document}